\pdfoutput=1
\documentclass{article}
\usepackage{arxiv}
\usepackage[utf8]{inputenc} % allow utf-8 input
\usepackage[T1]{fontenc}    % use 8-bit T1 fonts
\usepackage{hyperref}       % hyperlinks
\usepackage{url}            % simple URL typesetting
\usepackage{booktabs}       % professional-quality tables
\usepackage{amsfonts}       % blackboard math symbols
\usepackage{nicefrac}       % compact symbols for 1/2, etc.
\usepackage{microtype}      % microtypography
\usepackage{lipsum}		% Can be removed after putting your text content
\usepackage{graphicx}
\usepackage{natbib}
\usepackage{makecell}
\usepackage{multirow}
\usepackage{doi}
\usepackage{amsmath}

\title{GraMuFeN: Graph-based Multi-modal Fake News Detection in Social Media}

\author{ {\hspace{1mm}Makan Kananian} \\
	Department of Computer Engineering\\
	Sharif University of Technology,\\
        International Campus \\
	Kish Island, Iran \\
	\texttt{makankananian@gmail.com} \\
	\And
	{\hspace{1mm}Faitma Badiei} \\
	Department of Computer Engineering\\
	Sharif University of Technology, \\
 International Campus\\
	Kish Island, Iran \\
	\texttt{fatimabadiei78@gmail.com} \\
        \And
	{\hspace{1mm}S. AmirAli Gh. Ghahramnai} \\
	Department of Computer Engineering\\
	Sharif University of Technology\\
        International Campus\\
	Kish Island, Iran \\
	\texttt{ghahramani@sharif.edu} \\
}

\hypersetup{
pdftitle={GraMuFeN: Graph-based Multi-modal Fake News Detection in Social Media},
pdfsubject={q-bio.NC, q-bio.QM},
pdfauthor={Makan Kananian, Fatima Badiei, Dr. Amir Ali ghahramani},
pdfkeywords={Fake news detection, Social media, Graph Convolutional Neural Networks, Convolutional Neural Networks, Graph Neural Networks, GCN, CNN,  ResNet-152, Multi-modal, Text classification.},
}

\begin{document}
\maketitle
\begin{abstract}

The proliferation of social media platforms such as Twitter, Instagram, and Weibo has significantly enhanced the dissemination of false information. This phenomenon grants both individuals and governmental entities the ability to shape public opinions, highlighting the need for deploying effective detection methods. In this paper, we propose GraMuFeN, a model designed to detect fake content by analyzing both the textual and image content of news. GraMuFeN comprises two primary components: a text encoder and an image encoder. For textual analysis, GraMuFeN treats each text as a graph and employs a Graph Convolutional Neural Network (GCN) as the text encoder. Additionally, the pre-trained ResNet-152, as a Convolutional Neural Network (CNN), has been utilized as the image encoder. By integrating the outputs from these two encoders and implementing a contrastive similarity loss function, GraMuFeN achieves remarkable results. Extensive evaluations conducted on two publicly available benchmark datasets for social media news indicate a 10 \% increase in micro F1-Score, signifying improvement over existing state-of-the-art models. These findings underscore the effectiveness of combining GCN and CNN models for detecting fake news in multi-modal data, all while minimizing the additional computational burden imposed by model parameters.
\end{abstract}

% keywords can be removed
% \keywords{First keyword \and Second keyword \and More}

\section{Introduction}
In the current landscape of information dissemination, the proliferation of social media platforms, including Twitter, Instagram, Weibo, and others, has resulted in an era of interconnectedness and instantaneous communication. This digital revolution has not only facilitated global connectivity but has also inadvertently given rise to a formidable challenge - the rapid propagation of fake news \cite{wessel2016emergence}. 

In sharp contrast to the controlled environments of traditional media, social media's dynamic nature accelerates the spread of fake information, rendering the dissemination of fake news more pervasive than ever before. This concerning phenomenon has led to the potential manipulation of public sentiments by both malicious entities and governmental actors \cite{woolley2018computational} \cite{kalsnes2018fake}, thereby compromising the authenticity of the information.

Given this urgent situation, identifying and addressing fake news has become an important concern for researchers, experts, and decision-makers. While traditional methods for spotting fake news mostly focused on analyzing text, the diverse aspects of social media now require new approaches that can comprehend the complex interplay between text, images, and various forms of media. 

In our research, we tackle the challenge of detecting fake news in a multi-modal context, where news articles comprise both textual and image-based content. To address this complex problem, we propose the Graph-based Multi-modal Fake News Detection model (GraMuFeN). Our approach involves the integration of text and image encoders, utilizing the power of three distinct neural network architectures: Long Short Term Memory (LSTM), Graph Convolutional Neural Network (GCN), and Convolutional Neural Network (CNN).

Within the Text Encoder component, each textual input is treated as both sequential data and a graph-like structure, enabling the combination of LSTM and GCN networks. This enables us to capture complex relationships within the textual content effectively. Simultaneously, the Image Encoder utilizes the pre-trained ResNet-152 \cite{he2016deep} as the CNN model, enabling the extraction of features from the visual elements contained within each news. 

Subsequently, the feature vectors extracted by the Text and Image Encoders were combined to predict the label associated with each news by minimizing the cross entropy loss between the predictions and the ground-truth labels. At the same time and by employing a contrastive similarity loss, the GraMuFeN is trained to produce similar feature vectors for each text and its corresponding image. This dual training process enhances the model's ability to align text and image representations effectively while simultaneously improving its capacity for accurate label prediction. As a result, we achieved a 10 percent enhancement in the F1-score micro for multi-modal fake news detection on well-established benchmark datasets like Twitter \cite{boididou2015verifying}.

The rest of this paper is organized as follows. In Section \ref{Literature Review}, we present a brief review of the methods for spotting fake news in social media. We formally define the problem that is going to be solved in this research in Section \ref{Problem statement}. Section \ref{Proposed Method} provides the details related to the Text Encoder, the Image Encoder, the contrastive similarity loss, and the classifier used in our method. We describe the characteristics of the datasets used in this paper in Section \ref{Datasets}. Our extensive evaluations and their outcomes are summarized in Section \ref{Results}. Finally, we conclude the paper in Section \ref{Conclusion}.
\section{Literature Review}
\label{Literature Review}

% \subsection{Fake News Detection}
Following the guidance of key studies \cite{ruchansky2017csi} \cite{shu2017fake}, we describe 'fake news' as stories that are intentionally made up and can be proven wrong with real facts. Identifying fake news is not easy because it involves analyzing different aspects of the story including the message it is trying to convey, its place in society, and the images used. Research into detecting fake news is a field that is constantly growing, with many different methods being developed to tackle it.

In the early stages of fake news detection, most research focused on using single-mode data, mainly text.  \textit{Wu et al.}, used a special kind of system called a graph kernel-based hybrid support vector machine to understand how news spreads and to spot fake news \cite{wu2015false}. Some researchers looked at the way language was used in tweets to figure out if the news was fake \cite{shu2017fake}, while others used structural and cognitive features to detect fake news on social networks \cite{kwon2013prominent}.
But these early methods had problems. They were only good for specific topics and relied too much on manually picking out features, which could lead to biased results \cite{shu2017fake,kwon2013prominent}.

Later on, researchers started using other kinds of data like images from social media to find fake news \cite{ping2013review, gupta2012evaluating}. The use of deep learning models also started, improving the accuracy of fake news detection \cite{jin2017multimodal}. \textit{Steinebach et al.,} delve into the problem of fake images accompanying fake news \cite{steinebach2019fake}. They highlight that these manipulated images, often photo montages spliced from several images, are designed to make the fake news appear more authentic. To combat this, they developed a concept based on feature detection, indexing these features using a nearest-neighbor algorithm. This allows for the rapid comparison of a large number of images, identifying montages even if they have been manipulated in various ways.

In the paper written by \textit{Meng et al.} \cite{meng2022traceable}, the authors propose a proactive strategy to counteract fake news using traceable and authenticable image tagging. This strategy leverages a Decoupled Invertible Neural Network (DINN) to embed dual tags into news images before publication. However, the approach is not without its limitations. One significant concern is its generalization to unseen manipulations; the current setup might struggle to handle new types of manipulations that were not included in the training dataset. This indicates a potential area for further development to enhance the robustness of the system against a broader array of manipulations.
Even though now we are moving towards using multiple types of data, these early single-type data methods helped build the base for today's more advanced fake news detection techniques.

% \subsubsection{Multiple Modality}

As research progressed, it became clear that using different types of data together could make fake news detection more accurate. For example, some systems were designed to answer questions using deep learning networks \cite{xi2020visual}. Others combined text and images to better detect fake news \cite{farajtabar2017fake}.

A big step forward was the creation of EANN \cite{wang2018eann}, Unlike traditional methods that often rely on event-specific features, EANN is designed to identify fake news across various events, making it adaptable to new and unexpected news topics. It achieves this through a combination of multi-modal feature extraction and an event discriminator, ensuring both textual and visual content are analyzed while filtering out event-specific biases. This approach enhances its versatility in the ever-changing landscape of social media news. Following this, other powerful models like MVAE \cite{khattar2019mvae} emerged. MVAE is a special model designed to understand and analyze both these types of content together. It works by creating a combined representation of the text and images in news articles. This combined understanding helps MVAE decide if an article is fake or real. We should also mention SpotFake \cite{singhal2019spotfake}, which is designed to detect fake news by looking at both the words and pictures in an article. It uses transformers like BERT for understanding the text and convolutional neural networks like VGG-19 for analyzing images.

New methods kept coming, with CARMN \cite{song2021multimodal}. The special way that the CARMN model uses to look at both words and pictures in news stories is called "Crossmodal Attention Residual." This technique allows the model to focus on how words and pictures relate to each other. Additionally, this model uses "Multichannel convolutional neural Networks" to process and understand the information deeply. In simpler terms, "Crossmodal Attention Residual" helps CARMN pay special attention to the connections between text and images, while "Multichannel convolutional neural Networks" allow it to analyze the content from different perspectives or channels. This combination makes CARMN effective at detecting fake news. 

A significant advancement was AMFB \cite{kumari2021amfb} which stands out because it understands how image and text intertwine. The heart of AMFB lies in its "Attention-based multimodal Factorized Bilinear Pooling" technique. This means that AMFB meticulously focuses on the most crucial parts of the words and pictures, discerning which elements are vital for determining the authenticity of the news. It adeptly processes multiple types of information simultaneously, such as text and visuals. Furthermore, instead of merely merging the information from words and pictures, AMFB employs a sophisticated method of blending them, capturing their complex relationship. This approach empowers AMFB to discriminate between real and fake news stories with high accuracy.

Another model called FNR \cite{ghorbanpour2023fnr} was developed which evaluates how closely the content of a news story (both text and images) matches with known genuine news. By gauging this similarity, it gets an initial sense of the story's authenticity, FNR uses transformers (Bert and ViT (vision transformer) to deeply analyze the content, capturing intricate patterns.

In the paper written by \textit{Zhou et al. } \cite{zhou2020safe}, the authors introduce the Similarity-Aware Fake news detection (SAFE) method. The technique focuses on the identification of mismatches between text and images in news content, utilizing a multi-modal neural network to analyze the relationship between textual and visual elements and identify inconsistencies that are indicative of fake news.

% In conclusion, using multiple types of data together has greatly improved the accuracy of fake news detection. As researchers keep finding new ways to analyze data, we can hope for even better Models in the future.

\subsection{Comparison with Existing Methods}

In this study, we present the GraMuFeN, a model that leverages the power of Graph Convolutional Networks for text and ResNet-152 for images. 

Table \ref{Table:related-work} compares our proposed model with existing state-of-the-art models, including EANN, MVAE, SpotFake, CARMN, AMFB, and FNR, in terms of the type of text encoder, image encoder and the fusion used to combine textual and image features.

\begin{table}[!h]
\centering
\caption{Comparison of GraMuFeN with state-of-the-art models}
\label{Table:related-work}
\begin{tabular}{|c|c|c|c|} 
\hline
\textbf{Method}                                                & \textbf{Text Encoder}                                                      & \textbf{Image Encoder}                                            & \textbf{Fusion Type}                                                       \\ 
\hline
\makecell{EANN\\ \cite{wang2018eann}}                                                    & Text-CNN                                                                   & VGG19                                                             & Concat                                                                     \\ 
\hline
\makecell{MVAE\\ \cite{khattar2019mvae}}                                                        & BiLSTM                                                                     & VGG19                                                             & auto-encoder                                                               \\ 
\hline
\makecell{SpotFake\\ \cite{singhal2019spotfake}}                                                       & BERT                                                                       & VGG19                                                             & Concat                                                                     \\ 
\hline
\makecell{CARMN\\ \cite{song2021multimodal}}                                                         & \makecell{Word level\\sentence embeddings} & VGG19                                                             & \makecell{Concat +\\Crossmodal attention}  \\ 
\hline
\makecell{AMFB\\ \cite{kumari2021amfb}}                                                       & \makecell{Attention-based\\BiLSTM}           & \makecell{Attention-based\\CNN-RNN} & \makecell{Element wise\\multiplication}      \\ 
\hline
\makecell{FNR\\ \cite{ghorbanpour2023fnr}}                                                       & BERT                                                                       & \makecell{Visual\\transformer (ViT)} & \makecell{Concat +\\Similarity}              \\ 
\hline
GraMuFeN (Our Model) & GCN-LSTM                                                                   & \makecell{CNN\\(resnet-152)}        & \makecell{Concat +\\Similarity}              \\
\hline
\end{tabular}
\end{table}

\section{Problem statement}
\label{Problem statement}

Both of our datasets (Twitter and Weibo) contain some text that has been posted by a user with one or several images corresponding to that text. Each text $T$ has an image $T_{I}$ and a label $T_{L}$. For the same events with different texts, the image could be the same, especially in the Twitter dataset. Our model aims to use $T$ and $T_{I}$ to predict $T_{L}$, whether it is Fake or Real. In the next section, we delve deeper into our model.

\section{Proposed Method}
\label{Proposed Method}
\begin{figure}[h]
\begin{center}
\includegraphics[trim=1cm 0.5cm 0cm 0cm, scale = 0.5]{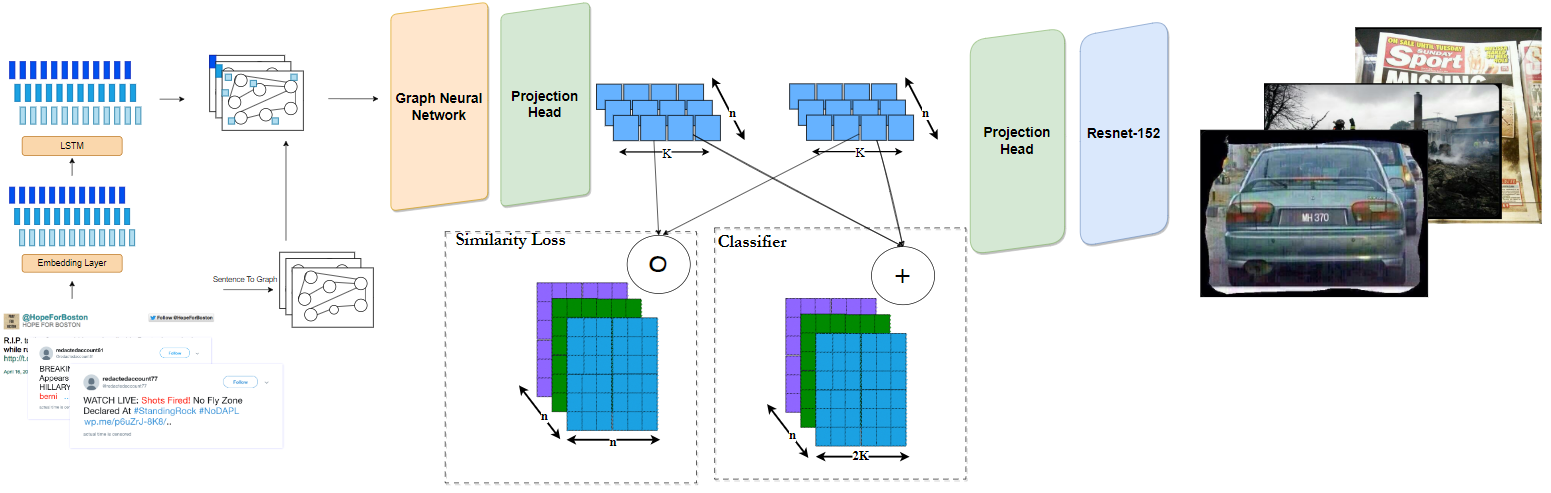}
\end{center}
\caption{Architecture of GraMuFeN}
\label{Figure: model architecture}
\end{figure}

The architecture of GraMuFeN consists of three main components. The Text Encoder, employs a sequence of multiple SageConv layers, generating embeddings for the provided textual content. The second component, the Image Encoder, employs the pre-trained ResNet-152 model to create embeddings corresponding to the images accompanying each textual instance. Ultimately, the text and image embeddings are combined and given to the third component, the classifier, which undertakes the responsibility of predicting the label of the news item. At the same time, the text and image embeddings are used to compute a supervised contrastive similarity loss \cite{khosla2020supervised}.  In Figure \ref{Figure: model architecture}, the architecture of our model is presented. In the subsequent sections, we explain each of the mentioned components in detail.

\subsection{The Text Encoder}
\begin{figure}[h]
\begin{center}
\includegraphics[trim=1cm 0.5cm 0cm 0cm, scale = 0.5]{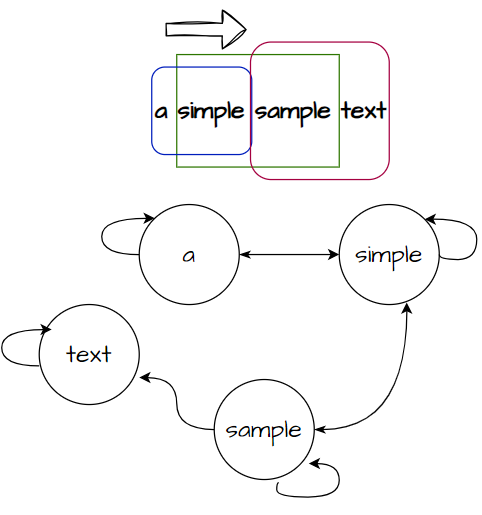}
\end{center}
\caption{Converting Text to Graph}
\label{Figure: sent2graph}
\end{figure}

The initial phase entails the transformation of each text (or sentence) into a graph. For this purpose, each word within a sentence constitutes an individual node, even instances of duplicate words. The establishment of connections between nodes is achieved through the utilization of a fixed-size context window. An example is provided in Figure \ref{Figure: sent2graph}. In this example, the context window size is 2. This context window slides through the sentence, forging links between nodes corresponding to words co-occurring within the window. Thus, the underlying graph structure for each sentence is derived. Moreover, to account for the requirements of GCN layers that will be applied later, we add self-loops for each node of the resulting graph. This procedure is carried out by the \textit{SentenceToGraph} module, as shown in Figure \ref{Figure: model architecture}. It is noteworthy, however, that at this time the node's features within the graph remain unspecified.

Following the creation of a distinct graph corresponding to each text (or sentence), the text is fed to an embedding layer.
The embedding layer's output is then fed to an LSTM network creating hidden states for each word in the sentence. At this time, these hidden states serve as the feature vectors for the respective nodes in the corresponding graph.
So far, we've successfully transformed each sentence into a graph, where the words in the sentence become nodes in the graph, and the LSTM-generated hidden states become the nodes' features. Next, we input this graph into our GCN (Graph Convolutional Neural Network) layers. Specifically, we utilize SageConv \cite{hamilton2017inductive} as GCN layers in GraMuFeN. Our model consists of three GCN layers, each incorporating a ReLU activation function to introduce non-linearity, enabling us to capture greater complexity.

To aggregate information from the GCN layers, we employ Global Mean Pooling, an aggregation technique that calculates the mean (average) of nodes' feature vectors for each graph, as follows:
    
    \begin{equation}
    \text{Global Mean Pooling}(X) = \frac{1}{N} \sum_{i=1}^{N} X_i
    \end{equation}
    
Here, \(X\) represents the feature vectors of all nodes in a graph, \(N\) is the number of nodes in the graph, and \(X_i\) represents the feature vector of the \(i\)-th node. The result is a single feature vector representing the aggregated information for each graph.

\begin{figure}[h]
\begin{center}
\includegraphics[trim=1cm 0cm 0 0cm, scale=0.45]{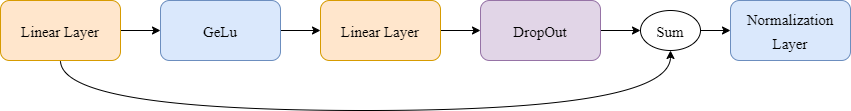}
\end{center}
\caption{Projection Head}
\label{Figure: Projection Head}
\end{figure}

After obtaining the embeddings from the GCN,  we employ a projection head \cite{ghorbanpour2023fnr} to map these embeddings to a new representational space: 
\begin{equation}
\begin{split}
z_{text} = \omega_{2} \times (\textit{gelu}(\omega_{1} \times \textit{X} + b_{1})) + (\omega_{1} \times \textit{X} + b_{1}) + b_{2}
\end{split}
\end{equation}
Where \(w_{1}\), \(w_{2}\), \(b_{1}\) and \(b_{2}\) are weights and
biases of linear layers inside the text projector and \textit{X} is the output of the Global Mean pool. The structure of this projection head is illustrated in Figure \ref{Figure: Projection Head}. $z_{text}$ is the final embedding for texts with the shape of $(B, e_{text})$ where $B$ is the batch size and $e_{text}$ is the size of the final embedding in the text projection head.

%-------FIGURE------%
%----------------------%

\subsection{Image Encoder}
As the Image Encoder, we employ the pre-trained ResNet-152 network.
This network takes batches of images in the shape of (B, C, W, H) representing the batch size, number of image channels,  width, and height of input images. ResNet-152 network projects the inputs into vectors of shape (B, 2048), where B is the batch size and 2048 is the size of the last linear layer in ResNet-152. Similar to the Text Encoder, we employ a projection head to map the outputs of the Image Encoder into a new space:
\begin{equation}
\begin{split}
z_{img} = \omega_{4} \times (\textit{gelu}(\omega_{3} \times \textit{V} + b_{3})) + (\omega_{3} \times \textit{V} + b_{3}) + b_{4}
\end{split}
\end{equation}
Where \(w_{3}\), \(w_{4}\), \(b_{3}\) and \(b_{4}\) are weights and
biases of linear layers inside the image projector. The structure of the image projection head is identical to that of the Text Encoder as illustrated in Figure \ref{Figure: Projection Head}. After applying the image projector, we obtain \(z_{img}\), which is the final embedding for images with the shape of $(B, e_{img})$ where $B$ is the batch size and $e_{img}$ is the size of the final embedding in the image projection head. 

\subsection{Similarity Loss}
To train the proposed model, we use a supervised contrastive similarity loss \cite{khosla2020supervised}. 
As explained previously, $z_{img}$ and $z_{text}$ are the final embeddings for images and texts with the shape of $(B, e_{img})$ and $(B, e_{text})$ for images and texts, where $e_{img}$ and $e_{text}$ are vector size of image and text embedding respectively. To obtain the level of similarity between these embeddings, their inner product is calculated:
\begin{equation}
\begin{split}
P = z_{text}z_{img}^{T}
\end{split}
\end{equation}
Here, we consider $P$ as the prediction matrix with the shape of $(B, B)$.

The contrastive loss function tries to maximize the similarity of each text and image to itself. To this aim and by means of inner product, we define the expected matrix as the average similarity of text-to-text and image-to-image based on the following formula \cite{radford2021learning, salama2021}:

% This loss function considers an image and a text to be the most similar to itself. Thus, we consider the expected matrix as the average similarity of the text-to-text and the image-to-image according to the following formula \cite{salama}:

\begin{equation}
\begin{split}
E = \textit{softmax}(\frac{z_{img}z_{img}^{T} + z_{text}z_{text}^{T}} {2})
\end{split}
\end{equation}

After calculating the $E$ matrix, we use cross-entropy to find the similarity loss. The contrastive similarity loss is the average of the text similarity loss \(l_{text}\), and the image similarity loss \(l_{img}\) \cite{radford2021learning}:

\begin{equation}
\begin{split}
l_{\text{text}} = -(E * \textit{log}(P) + (1 - E) * \textit{log}(1 - P))
\end{split}
\end{equation}
\begin{equation}
\begin{split}
l_{img} = -(E^{T} * \textit{log}(P^{T}) + (1 - E^{T}) * \textit{log}(1 - P^{T}))
\end{split}
\end{equation}
\begin{equation}
\begin{split}
l_{s} = \frac{l_{text} + l_{img}}{2}
\end{split}
\end{equation}

\subsection{The Classifier}
We concatenate the final text ($z_{text}$) and image ($z_{img})$ embeddings to create the combined embedding vector as follows \cite{ghorbanpour2023fnr}:

\begin{equation}
\begin{split}
z_{\text{combined}} = \text{Concat}(z_{\text{gcn}}, z_{\text{img}})
\end{split}
\end{equation}

\begin{figure}[!h]
\begin{center}
\includegraphics[trim=0cm 1cm 0 0cm, scale=0.45]{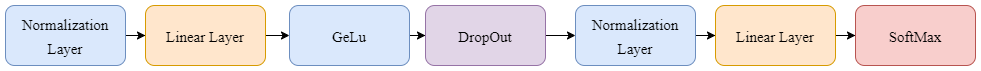}
\end{center}
\caption{Classifier}
\label{Figure: classifier}
\end{figure}

This embedding vector is then passed through two linear layers for fake news classification. After this linear mapping, we consider a vector of size (B, 2) with two classes. According to Figure \ref{Figure: classifier} and assuming \(w_{6}\), \(w_{5}\), \(b_{4}\) and \(b_{5}\) are weights and biases of linear layers inside the classifier, the classifier output ($Z$) can be defined as follows:

\begin{equation}
\begin{split}
Z &= \text{softmax}\left(\omega_{6} \times \left(\text{gelu}\left(\omega_{5} \times z_{\text{combined}} + b_{5}\right) + b_{6}\right)\right)
\end{split}
\end{equation}

%-------FIGURE------%
%----------------------%

After passing the vector through the softmax function, the classification loss can be stated as follows:

\begin{equation}
\begin{split}
l_{c} = -(\textit{L} \times \textit{log}(Z) + (1 - \textit{L}) \times \textit{log}(1 - Z))
\end{split}
\end{equation}

% (\(\alpha\) is the balancing factor for the weighted loss function):
% \begin{equation}
% \begin{split}
% l_{c} = -(\alpha \textit{L} \times \textit{log}(Z) + (1 - \textit{L}) \times \textit{log}(1 - Z))
% \end{split}
% \end{equation}
The final total loss is obtained by combining the classification loss and the similarity loss as follows:

\begin{equation}
\begin{split}
Loss = l_{c} + l_{s}
\end{split}
\end{equation}
% where $\lambda$ is the weight of the similarity loss in the final combination.

\section{Datasets}
\label{Datasets}
In this section, we elaborate on datasets that we used to benchmark our model, GraMuFeN. To this aim, we first begin with a description of the datasets used in this study. Then we explain the pre-processing employed for handling text and image data. 
% Finally, we discuss the results.

For the sake of evaluation, we choose two major datasets in this study:

\begin{itemize}
    \item Twitter dataset
    \item Weibo dataset
\end{itemize}

\subsection{Twiter Dataset}
The Twitter dataset is from MediaEval Verifying Multimedia Use
benchmark \cite{boididou2015verifying}, which is used for detecting fake content on Twitter. This dataset has two parts: the train set and the test set. The tweets in this dataset contain text content, attached images/videos, and additional social context information. In this work, we focus on detecting fake news by incorporating both text and image information. Thus, we remove the tweets without any text or image. We translated the whole dataset into English language using the Google Translate package For these two sets, there are no overlapping events among them. 
We further split the training dataset into training and validation sets, with the validation set comprising \textit{20 percent} of our training dataset. The important characteristics of this dataset are reported in Table \ref{Table: Twitter info}

In previous studies, authors who worked with this Twitter dataset, like us, performed several pre-processing steps, including the removal of links, duplicates, and more. Upon examining the dataset, we discovered that there are even more duplicates when tweets containing retweets (indicated by "RT" or "rt" in their text) and mentions of other users (e.g., "@sample-user") are removed. Previous works reported that the Twitter dataset comprises 12,237 tweets. However, after thorough pre-processing and the removal of all duplicates, this number was reduced to 11,817 tweets.

\begin{table}[!h]
\centering
\caption{Twitter Dataset Information}
\label{Table: Twitter info}
\begin{tabular}{|c|c|c|} 
\hline
\textbf{Dataset}       & \textbf{Label} & \textbf{Count}  \\ 
\hline
\multirow{2}{*}{Train} & Fake           & 6649            \\ 
\cline{2-3}
                       & Real           & 4599            \\ 
\hline
\multirow{2}{*}{Test}  & Fake           & 545             \\ 
\cline{2-3}
                       & Real           & 444             \\ 
\hline
\multicolumn{2}{|c|}{All Data}          & 12237           \\
\hline
\end{tabular}
\end{table}

\begin{figure}[!h]
\begin{center}
\includegraphics[trim=2cm 1cm 0 -0.5cm, scale=0.5]{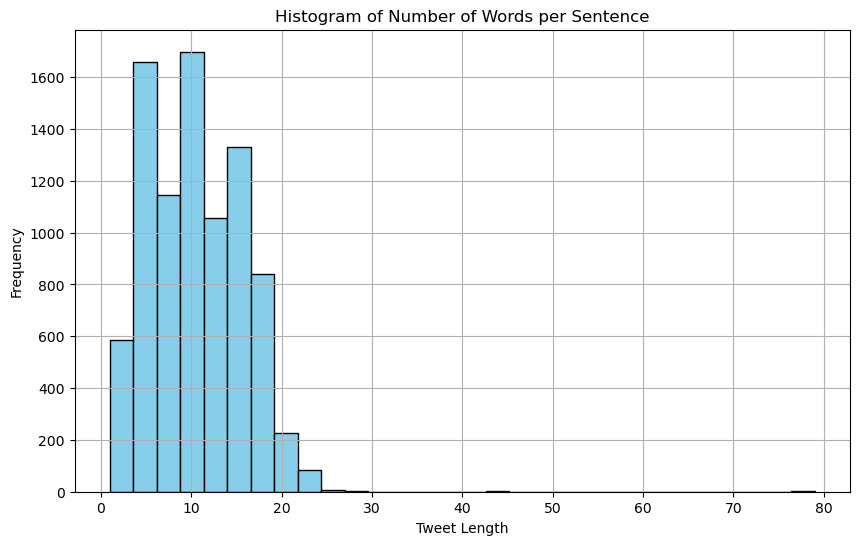}
\end{center}
\caption{Tweet Length Distribution in Twitter Dataset}
\label{Figure: Wrods count Twitter Datasets}
\end{figure}

We also analyzed the length of tweets in the Twitter dataset. The results are illustrated in Figure \ref{Figure: Wrods count Twitter Datasets}. Notably, the length of tweets in the Twitter dataset is shorter compared to the Weibo dataset with an average length of 10 words.

%-------FIGURE------%
%----------------------%

\subsection{Weibo Dataset}
This dataset \cite{jin2017multimodal} was collected from Weibo social media from 2012 to 2016 and is written in the Chinese language. Each line of data in this dataset also contains text, user information, and an image. The app’s authentication system has tagged texts. This database is divided by \cite{wang2018eann} \footnote{https://github.com/yaqingwang/EANN-KDD18} into three sets of train, validation, and test data, as the news events of each set are different. We utilized the same Datasets. We translated the whole dataset into English language using the Google Translate package For these two sets. Only the data with text and images are used. Moreover, in the Weibo dataset, each text (tweet) was linked with a different number of images. In this study, we considered only the first image linked to each text.  Table \ref{Table: Weibo info} lists the number of train/test data and fake/real data of the Weibo dataset.

\begin{table}[!h]
\centering
\caption{Weibo Dataset Information}
\label{Table: Weibo info}
\begin{tabular}{|c|c|c|} 
\hline
\textbf{Dataset}       & \textbf{Label} & \textbf{Count}  \\ 
\hline
\multirow{2}{*}{Train} & Fake           & 3748            \\ 
\cline{2-3}
                       & Real           & 3758            \\ 
\hline
\multirow{2}{*}{Test}  & Fake           & 999             \\ 
\cline{2-3}
                       & Real           & 995             \\ 
\hline
\multicolumn{2}{|c|}{All Data}          & 9500            \\
\hline
\end{tabular}
\end{table}

%-------FIGURE------%
\begin{figure}[!h]
\begin{center}
\includegraphics[trim=2cm 1cm 0 -0.5cm, scale=0.5]{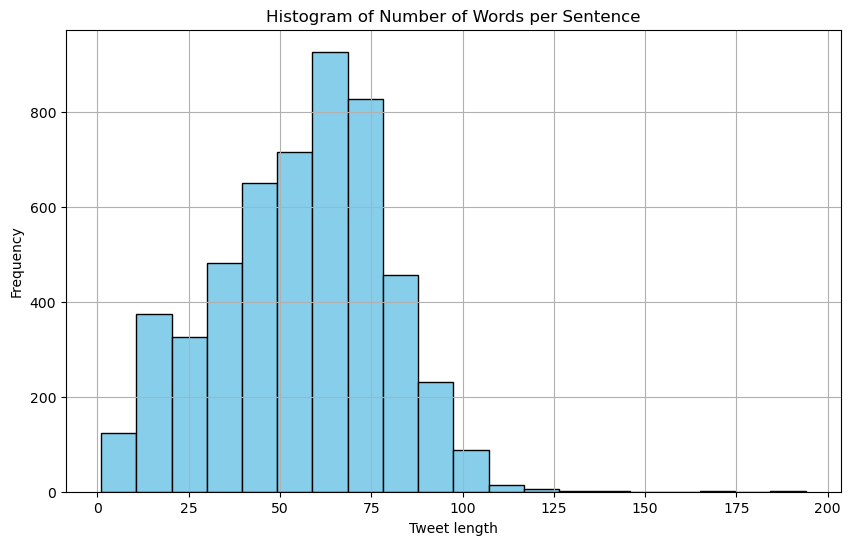}
\end{center}
\caption{Tweet length Distribution in Weibo Dataset}
\label{Figure: Wrods count WEibo Datasets}
\end{figure}
%----------------------%
We also analyzed the length of tweets in the Weibo dataset. The results are presented in Figure \ref{Figure: Wrods count WEibo Datasets}. As evident, the sequence length in the Weibo dataset is significantly greater in comparison to the Twitter dataset.
\section{Results}
\label{Results}
We utilized PyTorch version 2.0.1 with CUDA 11.8 (cu118) support and Torch Geometrics for the development and training of our proposed model. The experiments were conducted on the Google Colab Pro platform, leveraging a T4 GPU with approximately 15GB of VRAM, 16GB of RAM, and 30GB of storage. 
\subsection{Results in Twitter Dataset}

For the Twitter dataset, we first pre-train the text encoder and image encoder separately on texts and images, respectively. Then we combine the pre-trained encoders resulting in the combined model and we start to finetune it. In the following sections, we elaborate on these training procedures.
\subsubsection{Pre-training Text Encoder}
We trained the Text Encoder separately on the text data in the Twitter dataset. We employed a  pre-trained embedding layer, specifically the Google News Word2Vec embedding \cite{mikolov2013efficient}. This embedding layer has been trained on extensive text corpora and provides embeddings for 3,000,000 words, each consisting of a 300-dimensional feature vector. It's worth noting that we have frozen all the pre-trained embeddings. Consequently, this layer generates a 300-dimensional feature vector for each word. The necessary information for the pretraining procedure is reported in Table \ref{tab: text_twitter_pretrain} .
\begin{table}[!h]
		\begin{center}
			\caption{The Text Encoder Pretraining for Twitter Dataset}
                \label{tab: text_twitter_pretrain}
			\begin{tabular}{ p{4.5cm} |p{2cm}}
				\noalign{\hrule height 1.5pt}
				\textbf{Initial learning rate} &  3e-3 \\
				\hline
				\textbf{batch size}&256\\
				\hline
				\textbf{Epoch} & 30\\
				\hline
				\textbf{Optimizer} & AdamW\\
				
				\noalign{\hrule height 1.5pt}				
			\end{tabular}
		\end{center}
	\end{table}

Furthermore, we incorporated a learning rate scheduler, \textit{ReduceLROnPlateau}, with a factor of 0.8 and a patience parameter of 3. Additionally, to enhance training stability, we implemented gradient clipping, capping the gradients at a maximum norm of 1, We utilized Optuna for hyperparameter tuning. 

The configuration of the Text Encoder is presented in Table \ref{tab: twitter_text_config}.

\begin{table}[!h]
\caption{Twitter Text Encoder Configuration}
\label{tab: twitter_text_config}
\centering
\begin{tabular}{|c|c|c|} 
\hline
\multicolumn{3}{|c|}{\textbf{Text Encoder for Twitter }}                                                                                                        \\ 
\hline
\textbf{Layers}     & \textbf{Configs}                                                                                    & \textbf{Activation Function}  \\ 
\hline
Embedding           & \begin{tabular}[c]{@{}c@{}}n vocabulary: 3,000,000\\output dim: 300\end{tabular}                    &                               \\ 
\hline
LSTM                & \begin{tabular}[c]{@{}c@{}}input dim: 300\\n Layers: 3\\Dropout = 0.3\\hidden dim: 256\end{tabular} &                               \\ 
\hline
Sageconv            & \begin{tabular}[c]{@{}c@{}}input dim: 256\\n layers: 3\\L2-normalized\\hidden dim: 512\end{tabular} & Relu                          \\ 
\hline
Global Mean pool    & \begin{tabular}[c]{@{}c@{}}input dim:512\\output dim:512\end{tabular}                               &                               \\ 
\hline
Dropout             & Percentage: 50\%                                                                                    &                               \\ 
\hline
Linear (Classifier) & \begin{tabular}[c]{@{}c@{}}input dim: 512\\n Layers: 1\\output dim: 2\end{tabular}                  &                               \\
\hline
\end{tabular}
\end{table}

The performance of the pre-trained Text Encoder over the Twitter dataset is presented in Table \ref{tab:textencoder_pretrain_performance}. In this table, the proposed Text Encoder is named \textit{GCNLSTM}, and the performance of other models is presented based on what is reported in \cite{ghorbanpour2023fnr}.

\begin{table}[!h]
\caption{Pre-trained Text Encoder Performance on Twitter Dataset}
% \begin{adjustwidth}{-0.5cm}{}
\centering
\begin{tabular}{|c|c|c|c|c|c|c|c|c|c|} 
\hline
\multicolumn{2}{|c|}{\multirow{2}{*}{}}                                              & \multicolumn{2}{c|}{} & \multicolumn{3}{c|}{Fake}     & \multicolumn{3}{c|}{Real}      \\ 
\cline{3-10}
\multicolumn{2}{|c|}{}                                                               & Accuracy & F1-micro   & Precision & Recall & F1-score & Precision & Recall & F1-score  \\ 
\hline
\multirow{5}{*}{Text} & \begin{tabular}[c]{@{}c@{}}Logistic\\Regression\end{tabular} & 0.62     & 0.62       & 0.69      & 0.55   & 0.61     & 0.56      & 0.70   & 0.62      \\ 
\cline{2-10}
                      & SVM                                                          & 061      & 0.61       & 0.68      & 0.55   & 0.61     & 0.55      & 0.68   & 0.62      \\ 
\cline{2-10}
                      & BiLSTM                                                       & 0.61     & 0.60       & 0.62      & 0.73   & 0.67     & 0.58      & 0.45   & 0.51      \\ 
\cline{2-10}
                      & Bert                                                         & 0.69     & 0.64       & 0.67      & 0.68   & 0.68     & 0.60      & 0.59   & 0.59      \\ 
\cline{2-10}
                      & \begin{tabular}[c]{@{}c@{}}GCN\\LSTM\end{tabular}            & \textbf{0.70}     & \textbf{0.69}      & \textbf{0.70}      & \textbf{0.69}   & \textbf{0.69}     & \textbf{0.69}      & \textbf{0.70}   & \textbf{0.70}      \\
\hline
\end{tabular}

\label{tab:textencoder_pretrain_performance}
% \end{adjustwidth}
\end{table}
\subsubsection{Pre-training Image Encoder}
We also pre-trained our Image Encoder, Resnet-152, on Twitter images. The Resnet-152 has been trained on the Imagnet Dataset and has a powerful understanding of image embeddings.

Like our Text Encoder, we used a learning rate scheduler \textit{ReduceOnPlateau} with a factor of 0.7 and a patience parameter of 5. The necessary information for the pretraining procedure of the image encoder is reported in Table \ref{tab: image_twitter_pretrain}. 
\begin{table}[!h]
		\begin{center}
			\caption{The Image Encoder Pretraining for Twitter Dataset}
                \label{tab: image_twitter_pretrain}
			\begin{tabular}{ p{4.5cm} |p{2cm}}
				\noalign{\hrule height 1.5pt}
				\textbf{Initial learning rate} &  1e-4 \\
				\hline
				\textbf{batch size}&128\\
				\hline
				\textbf{Epoch} & 10\\
				\hline
				\textbf{Optimizer} & AdamW\\
    \hline
    \textbf{Weight Decay} & 0.07\\
				
				\noalign{\hrule height 1.5pt}				
			\end{tabular}
		\end{center}
	\end{table}

The performance of the pre-trained Image Encoder over the Twitter Image dataset is presented in Table \ref{tab:imageencoder_pretrain_performance}. In this table, our Image Encoder is named \textit{ResNet-152}.
\begin{table}[!h]
\caption{Pre-trained Image Encoder Performance on Twitter Dataset}
% \begin{adjustwidth}{-0.5cm}{}
\centering
\begin{tabular}{|c|c|c|c|c|c|c|c|c|c|}
\hline
\multicolumn{2}{|c|}{\multirow{2}{*}{}} & \multicolumn{2}{c|}{} & \multicolumn{3}{c|}{Fake} & \multicolumn{3}{c|}{Real} \\ 
\cline{3-10}
\multicolumn{2}{|c|}{} & Accuracy & F1-micro & Precision & Recall & F1-score & Precision & Recall & F1-score \\ 
\hline
\multirow{4}{*}{Image} & CNN & 0.62 & 0.62 & 0.69 & 0.55 & 0.61 & 0.56 & 0.55 & 0.61 \\ 
\cline{2-10}
 & VGG19 & 0.68 & 0.68 & 0.74 & 0.64 & 0.69 & 0.62 & \textbf{0.72} & 0.67 \\ 
\cline{2-10}
 & ViT & 0.72 & \textbf{0.76} & 0.74 & 0.86 & 0.80 & \textbf{0.79} & 0.63 & \textbf{0.7} \\ 
\cline{2-10}
 & \begin{tabular}[c]{@{}c@{}}Resnet\\152\end{tabular} & \textbf{0.74} & 0.74 & \textbf{0.74} & \textbf{0.91} & \textbf{0.82} & 0.73 & 0.44 & 0.55 \\
\hline
\end{tabular}
\label{tab:imageencoder_pretrain_performance}
% \end{adjustwidth}
\end{table}
\subsubsection{Multi-modal Results}
In this section, we report results related to our final model (GraMuFeN) that combines the pre-trained Text and Image encoders as what was illustrated previously in Figure \ref{Figure: model architecture}. It is important to note that the embedding layer of the Text Encoder has been frozen during the training procedure. Also, we incorporated a learning rate scheduler, ReduceLROnPlateau, with a factor of 0.9 and a patience parameter of 2. Additionally, to enhance training stability, we implemented gradient clipping, capping the gradients at a maximum norm of 1, plus Torch AMP Auto cast for mixed precision training. The necessary information related to the training procedure of this model is reported in Table \ref{tab: combined_training_setting}.
\begin{table}[!h]
		\begin{center}
			\caption{The Model Training Setting for Twitter Dataset}
                \label{tab: combined_training_setting}
			\begin{tabular}{ p{4.5cm} |p{2cm}}
				\noalign{\hrule height 1.5pt}
				\textbf{batch size} &  64 \\
				\hline
				\textbf{Epochs}&30\\
				\hline
				\textbf{Optimizer} & AdamW\\
				\hline
                    \textbf{Weight Decay} & 0.07\\
                    \hline
				\textbf{Learning Rate for Text Encoder} & 1e-5\\
    \hline
    \textbf{Learning Rate For Text Projection Head}&5e-3\\
    \hline
    \textbf{Learning Rate for Image Encoder}&1e-7\\
    \hline
    \textbf{Learning Rate for Image Projection Head}&5e-3\\
    \hline\textbf{Learning Rate for Classifier}& 5e-3\\
				
				\noalign{\hrule height 1.5pt}				
			\end{tabular}
		\end{center}
	\end{table}
 
\begin{figure}
\begin{center}
\includegraphics[trim=2cm 1cm 0 -0.5cm, scale=0.6]{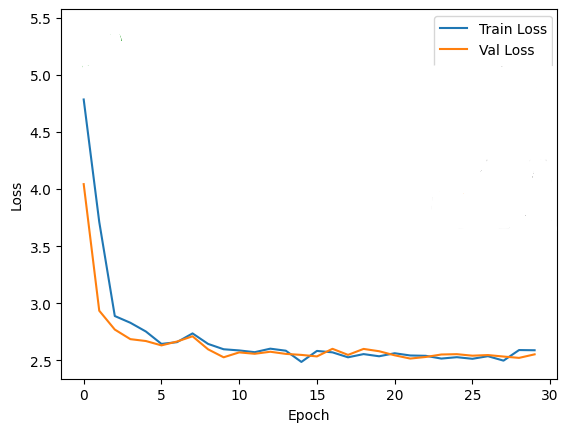}
\end{center}
\caption{Training and Validation loss in Twitter Dataset}
\label{Figure: Training and Validation loss Twitter}
\end{figure}

In Figure \ref{Figure: Training and Validation loss Twitter} the learning curve of training and validation is reported. Obviously, the training procedure is smooth and stable.
 
%-------FIGURE------%

We report the performance of our model (GraMuFeN) in Table \ref{tab: final twitter}. Results related to other models are presented based on reports in \cite{ghorbanpour2023fnr}. 

Regarding the findings presented in this table, the combination of the Text and Image Encoders has led to a noteworthy enhancement in accuracy, precision, recall, and the F1 score. Concerning accuracy, there is an observed advancement of 19 percent and 15 percent when compared to the Text Encoder (see Table \ref{tab: text_twitter_pretrain}) and the Image Encoder (see Table \ref{tab:imageencoder_pretrain_performance}), respectively. The joint utilization of Text and Image Encoders yielded an F1-micro score of 0.89, whereas the F1-micro scores for the Text Encoder and Image Encoder were 0.69 and 0.74, respectively. A similar pattern emerges for precision, recall, and the F1-Score pertaining to both fake and real samples within the dataset.

\begin{table}[!h]
% \begin{adjustwidth}{-1cm}{}
\centering
\caption{Performance of the GraMuFeN on Twitter Dataset}
\label{tab: final twitter}
\begin{tabular}{|c|c|c|c|c|c|c|c|c|c|} 
\hline
\multicolumn{2}{|c|}{\multirow{2}{*}{}}                                            & \multicolumn{2}{c|}{}         & \multicolumn{3}{c|}{Fake}                     & \multicolumn{3}{c|}{Real}                      \\ 
\cline{3-10}
\multicolumn{2}{|c|}{}                                                             & Accuracy      & F1-micro      & Precision     & Recall        & F1-score      & Precision     & Recall        & F1-score       \\ 
\hline
\multirow{7}{*}{\begin{tabular}[c]{@{}c@{}}Multi\\Model\end{tabular}} & EANN       & 0.69          & 0.69          & 0.75          & 0.58          & 0.65          & 0.62          & 0.76          & 0.69           \\ 
\cline{2-10}
                                                                      & MVAE       & 0.67          & 0.67          & 0.70          & 0.69          & 0.69          & 0.63          & 0.64          & 0.63           \\ 
\cline{2-10}
                                                                      & SpotFake   & 0.77          & 0.76          & 0.72          & \textbf{0.92} & 0.81          & 0.85          & 0.56          & 0.68           \\ 
\cline{2-10}
                                                                      & CARMN      & 0.73          & 0.73          & 0.70          & 0.88          & 0.78          & 0.78          & 0.54          & 0.64           \\ 
\cline{2-10}
                                                                      & AMFD       & 0.75          & 0.75          & 0.76          & 0.79          & 0.78          & 0.73          & 0.70          & 0.71           \\ 
\cline{2-10}
                                                                      & FNR-S      & 0.79          & 0.79          & 0.78          & 0.85          & 0.82          & 0.79          & 0.71          & 0.75           \\ 
\cline{2-10}
                                                                      & GraMuFeN & \textbf{0.89} & \textbf{0.89} & \textbf{0.92} & 0.85          & \textbf{0.89} & \textbf{0.86} & \textbf{0.93} & \textbf{0.89}  \\
\hline
\end{tabular}
% \end{adjustwidth}
\end{table}

When compared with other models in Table \ref{tab: final twitter}, GraMuFeN outperforms them across various performance metrics. Specifically, in regard to accuracy, our proposed model achieves an accuracy of 0.89, which is notably 10 percent higher than the most advanced state-of-the-art model (referred to as FNR-S). This same trend is evident for the F1-micro score as well. 

For instances of fake samples, GraMuFeN attains a precision of 0.92, surpassing the leading model by a substantial 14 percent. However, in terms of recall, the SpotFake model takes the lead. Conversely, when it comes to the F1-Score, GraMuFeN again emerges as the frontrunner, showcasing an improvement of over 7 percent in comparison to the best existing model.

For the category of real samples, a parallel pattern unfolds. Our model enhances performance metrics compared to the prevailing state-of-the-art models. Notably, GraMuFeN achieves increases of 7 percent, 17 percent, and 14 percent in terms of precision, recall, and F1-score, respectively, when matched against the existing state-of-the-art models.

\subsection{Results in Weibo Dataset}
In the case of the Weibo dataset, we began by pre-training the image encoder. In contrast to the Twitter dataset, we did not employ pre-trained embeddings for our Text Encoder. Instead, we allowed the embedding layer in the Text Encoder to train concurrently with our projection heads and classifier during our multi-modal training phase. We will thoroughly discuss the reasons for choosing this approach for the Weibo dataset. 
In the following sections, we will provide detailed explanations of these training procedures.

\subsubsection{Text Encoder Results}
In our approach for the Weibo dataset, we deliberately chose not to use pre-trained embeddings in the Text Encoder. The reason behind this decision is that, unlike the Twitter dataset, the Weibo dataset contains longer (as illustrated in Figure \ref{Figure: Wrods count WEibo Datasets}) and more meaningful sentences with a greater variety of words.
As such, we decided to train our own embedding layer on Weibo dataset from scratch. 

With the aim of showcasing the performance of the Text Encoder, we started to train it on the Weibo text dataset. The configuration of the Text Encoder is presented in Table \ref{Table: Weibo config}, We utilized Optuna for hyperparameter tuning.
Necessary information for using our Text Encoder during training is provided in Table \ref{tab: text_weibo}

\begin{table}[!h]
		\begin{center}
			\caption{The Image Encoder Pretraining for Weibo Dataset}
                \label{tab: text_weibo}
			\begin{tabular}{ p{4.5cm} |p{2cm}}
				\noalign{\hrule height 1.5pt}
				\textbf{Initial learning rate} &  5e-3 \\
				\hline
				\textbf{batch size}&32\\
				\hline
				\textbf{Epoch} & 50\\
				\hline
				\textbf{Optimizer} & AdamW\\
				
				\noalign{\hrule height 1.5pt}				
			\end{tabular}
		\end{center}
	\end{table}

Furthermore, we incorporated a learning rate scheduler, \textit{ReduceLROnPlateau}, with a factor of 0.8 and a patience parameter of 3. Additionally, to enhance training stability, we implemented gradient clipping, capping the gradients at a maximum norm of 1.

\begin{table}[!h]
% \begin{adjustwidth}{-0.5cm}{}
\centering
\caption{Weibo Text Encoder Configuration}
\label{Table: Weibo  config}
\begin{tabular}{|c|c|c|} 
\hline
\multicolumn{3}{|c|}{\textbf{Text Model Weibo}}                                                                                          \\ 
\hline
\textbf{Layers}     & \textbf{Configs}                                                                   & \textbf{Activation Function}  \\ 
\hline
Embedding           & \begin{tabular}[c]{@{}c@{}}n vocabulary: 8,991\\output dim: 16\end{tabular}        &                               \\ 
\hline
Dropout             & Percentage: 50\%                                                                   & \multicolumn{1}{l|}{}         \\ 
\hline
LSTM                & \begin{tabular}[c]{@{}c@{}}input dim: 16\\n Layers: 2\\hidden dim: 32\end{tabular} &                               \\ 
\hline
Sageconv            & \begin{tabular}[c]{@{}c@{}}input dim: 32\\n layers: 3\\hidden dim: 32\end{tabular} & Relu                          \\ 
\hline
Global Mean pool    & \begin{tabular}[c]{@{}c@{}}input dim:32\\output dim:32\end{tabular}                &                               \\ 
\hline
Dropout             & Percentage: 50\%                                                                   &                               \\ 
\hline
Linear (Classifier) & \begin{tabular}[c]{@{}c@{}}input dim: 32\\n Layers: 1\\output dim: 2\end{tabular}  &                               \\
\hline
\end{tabular}
% \end{adjustwidth}

\end{table}
The performance of our Text Encoder over the Weibo dataset is presented in Table \ref{tab:model-text-weibo}. In this table, the proposed Text Encoder is named GCNLSTM.

\begin{table}[!h]
% \begin{adjustwidth}{-0.5cm}{}

\centering
\caption{Text Encoder Performance on Weibo Dataset}
\label{tab:model-text-weibo}
\begin{tabular}{|c|c|c|c|c|c|c|c|c|c|} 
\hline
\multicolumn{2}{|c|}{\multirow{2}{*}{}}                                              & \multicolumn{2}{c|}{}         & \multicolumn{3}{c|}{Fake}                     & \multicolumn{3}{c|}{Real}                      \\ 
\cline{3-10}
\multicolumn{2}{|c|}{}                                                               & Accuracy      & F1-micro      & Precision     & Recall        & F1-score      & Precision     & Recall        & F1-score       \\ 
\hline
\multirow{5}{*}{Text} & \begin{tabular}[c]{@{}c@{}}Logistic\\Regression\end{tabular} & 0.71          & 0.71          & 0.71          & 0.80          & 0.75          & 0.71          & 0.59          & 0.65           \\ 
\cline{2-10}
                      & SVM                                                          & 0.70          & 0.70          & 0.72          & 0.74          & 0.73          & 0.67          & 0.65          & 0.66           \\ 
\cline{2-10}
                      & BiLSTM                                                       & 0.66          & 0.66          & 0.62          & 0.78          & 0.69          & 0.73          & 0.55          & 0.63           \\ 
\cline{2-10}
                      & Bert                                                         & \textbf{0.81} & \textbf{0.81} & \textbf{0.81} & \textbf{0.81} & \textbf{0.81} & 0.69          & \textbf{0.82} & \textbf{0.81}  \\ 
\cline{2-10}
                      & \begin{tabular}[c]{@{}c@{}}GCN\\LSTM\end{tabular}            & 0.73          & 0.73          & 0.73          & 0.76          & 0.75          & \textbf{0.74} & 0.70          & 0.72           \\
\hline
\end{tabular}
% \end{adjustwidth}
\end{table}

\subsubsection{Pre-training Image Encoder}
We did pre-train our Image Encoder, \textbf{Resnet-152}, on Weibo images. Like our Text Encoder, we used a learning rate scheduler \textit{ReduceOnPlateau} with a factor of 0.8 and a patience parameter of 5. Additionally, to enhance training stability, we implemented gradient clipping, capping the gradients at a maximum norm of 1. The necessary information for the pretraining procedure of the image encoder is reported in Table \ref{tab: image_weibo_pretrain}.

\begin{table}[!h]
		\begin{center}
			\caption{The Image Encoder Pretraining for Weibo Dataset}
                \label{tab: image_weibo_pretrain}
			\begin{tabular}{ p{4.5cm} |p{2cm}}
				\noalign{\hrule height 1.5pt}
				\textbf{Initial learning rate} &  1e-4 \\
				\hline
				\textbf{batch size}&64\\
				\hline
				\textbf{Epoch} & 30\\
				\hline
				\textbf{Optimizer} & AdamW\\
				\noalign{\hrule height 1.5pt}				
			\end{tabular}
		\end{center}
	\end{table}

The performance of the pre-trained Image Encoder over the Weibo Image dataset is presented in Table \ref{tab:model-image-weibo}. In this table, our Image Encoder is named \textit{ResNet-152}.

\begin{table}[!h]
% \begin{adjustwidth}{-0.3cm}{}
\centering
\caption{Pre-trained Image Encoder Performance on Weibo Dataset}
\label{tab:model-image-weibo}
\begin{tabular}{|c|c|c|c|c|c|c|c|c|c|} 
\hline
\multicolumn{2}{|c|}{\multirow{2}{*}{}}                                      & \multicolumn{2}{c|}{}         & \multicolumn{3}{c|}{Fake}                     & \multicolumn{3}{c|}{Real}                      \\ 
\cline{3-10}
\multicolumn{2}{|c|}{}                                                       & Accuracy      & F1-micro      & Precision     & Recall        & F1-score      & Precision     & Recall        & F1-score       \\ 
\hline
\multirow{4}{*}{Image} & CNN                                                 & 0.52          & 0.50          & \textbf{0.79} & 0.24          & 0.38          & 0.58          & 0.55          & 0.63           \\ 
\cline{2-10}
                       & VGG19                                               & 0.60          & 0.60          & 0.60          & 0.61          & 0.60          & 0.60          & \textbf{0.72} & 0.59           \\ 
\cline{2-10}
                       & ViT                                                 & 0.68          & 0.68          & 0.67          & 0.69          & 0.68          & \textbf{0.79} & 0.68          & 0.67           \\ 
\cline{2-10}
                       & \begin{tabular}[c]{@{}c@{}}Resnet\\152\end{tabular} & \textbf{0.71} & \textbf{0.71} & 0.72          & \textbf{0.72} & \textbf{0.72} & 0.70          & 0.70          & \textbf{0.70}  \\
\hline
\end{tabular}
% \end{adjustwidth}
\end{table}

\subsubsection{Multi-modal Results}
In this section, we report results related to our final model (GraMuFeN) that combines the Text and pre-trained Image encoder as what was illustrated previously in Figure \ref{Figure: model architecture}. It is important to note that the text encoder is not pre-trained and it will be trained from the ground up during this training procedure. Also, we incorporated a learning rate scheduler, ReduceLROnPlateau, with a factor of 0.9 and a patience parameter of 2 plus, Torch AMP Auto cast for mixed precision training. Additionally, to enhance training stability, we implemented gradient clipping, capping the gradients at a maximum norm of 1. The necessary information related to the training procedure of this model is reported in Table \ref{tab: combined_training_setting_weibo}.

\begin{table}[!h]
\centering
\caption{The Model Training Setting for Weibo Dataset}
\label{tab: combined_training_setting_weibo}
\begin{tabular}{l|l} 
\hline
\textbf{Batch size}                              & 30     \\ 
\hline
\textbf{Epochs}                                  & 20     \\ 
\hline
\textbf{Optimizer}                               & AdamW  \\ 
\hline
\textbf{Weight Decay for Image Encoder}          & 0.07   \\ 
\hline
\textbf{Weight Decay for Image Projection Head}  & 0.07   \\ 
\hline
\textbf{Learning Rate for Text Encoder}          & 5e-3   \\ 
\hline
\textbf{Learning Rate For Text Projection Head}  & 5e-3   \\ 
\hline
\textbf{Learning Rate for Image Encoder}         & 1e-7   \\ 
\hline
\textbf{Learning Rate for Image Projection Head} & 5e-3   \\ 
\hline
\textbf{Learning Rate for Classifier}            & 5e-3   \\
\hline
\end{tabular}
\end{table}

\begin{figure}[!h]
\begin{center}
\includegraphics[trim=2cm 1cm 0 -0.5cm, scale=0.6]{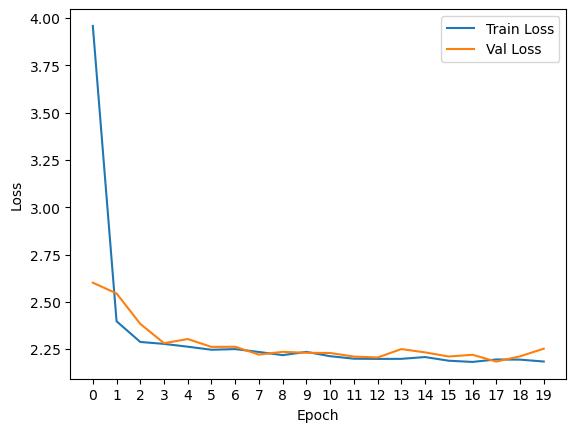}
\end{center}
\caption{Training and Validation loss in Weibo Dataset}
\label{Figure: Training and Validation loss Weibo}
\end{figure}

In Figure \ref{Figure: Training and Validation loss Weibo} the learning curve of training and validation is reported. Obviously, the training procedure is smooth and stable.
We report the performance of our proposed model in Table \ref{Table: final Weibo}. 
%-------FIGURE------%
%----------------------%

\begin{table}[!h]
% \begin{adjustwidth}{-1cm}{}
\centering
\caption{Performance of the GraMuFeN on Weibo Dataset}
\label{Table: final Weibo}
\begin{tabular}{|c|c|c|c|c|c|c|c|c|c|} 
\hline
\multicolumn{2}{|c|}{\multirow{2}{*}{}}                                            & \multicolumn{2}{c|}{}         & \multicolumn{3}{c|}{Fake}                     & \multicolumn{3}{c|}{Real}                      \\ 
\cline{3-10}
\multicolumn{2}{|c|}{}                                                             & Accuracy      & F1-micro      & Precision     & Recall        & F1-score      & Precision     & Recall        & F1-score       \\ 
\hline
\multirow{7}{*}{\begin{tabular}[c]{@{}c@{}}Multi\\Model\end{tabular}} & EANN       & 0.81          & 0.81          & 0.89          & 0.66          & 0.76          & 0.77          & 0.93          & 0.85           \\ 
\cline{2-10}
                                                                      & MVAE       & 0.79          & 0.79          & \textbf{0.89}          & 0.65          & 0.75          & 0.74          & \textbf{0.93} & 0.82           \\ 
\cline{2-10}
                                                                      & SpotFake   & 0.86          & 0.86          & 0.87          & 0.92          & \textbf{0.90} & 0.81          & 0.70          & 0.75           \\ 
\cline{2-10}
                                                                      & CARMN      & 0.84          & 0.85          & 0.86          & \textbf{0.93} & 0.89          & 0.81          & 0.66          & 0.73           \\ 
\cline{2-10}
                                                                      & AMFD       & 0.83          & 0.83          & 0.86          & 0.90          & 0.88          & 0.75          & 0.68          & 0.71           \\ 
\cline{2-10}
                                                                      & FNR-S      & 
                                                                      0.87 & 
                                                                      0.87 & 0.87 & 0.89          & 0.88          & 0.88          & 0.87          & \textbf{0.88} \\ 
\cline{2-10}
                                                                      & GraMuFeN & \textbf{0.87}          & \textbf{0.87}         & 0.86          & 0.89          & 0.87 & \textbf{0.88} & 0.84          & 0.86           \\
\hline
\end{tabular}
% \end{adjustwidth}
\end{table}

When compared to other models presented in Table \ref{Table: final Weibo}, our proposed approach outperforms all of them and has the same results as the FNR-S model, across various performance metrics.

We attribute the primary challenge encountered during the training of our model on the Weibo dataset to its length. Our proposed model can certainly outperform most state-of-the-art models except FNR-S, where we achieved the same results. There are several contributing factors. Notably, we did not impose a specific word limit on the text, whereas in \cite{ghorbanpour2023fnr}, texts were limited to 32 words on Twitter and 200 characters on Weibo. Overall results are the same. The slight score difference where we fell short could also be attributed to errors. The Weibo dataset is in the Chinese language, and the translation method, as well as data pre-processing before and after translation, can impact results significantly.

\section{Conclusion and Future Works}
\label{Conclusion}

In this paper, we introduced GraMuFeN, a model that leverages the capabilities of Graph Convolutional Neural Networks (GCN) and Long Short-Term Memory (LSTM) networks as the core of its text encoder, and Resnet-152 as its image encoder. Graph neural networks have demonstrated exemplary performance across Natural Language Processing (NLP) tasks. Our Vision was to illustrate the potential of Graph Neural Networks in the domain of Fake News Detection. As evidenced by our results, we achieved commendable outcomes with a model that is considerably smaller compared to BERT and ViT. Our model was benchmarked using two datasets, namely Twitter and Weibo, which have been utilized in previous analogous works. As delineated in our results section \ref{Results}, there is a noticeable enhancement in the F1 score and the overall performance of the model.
We believe that there exists further scope for enhancement. For example, incorporating more contextual features to the model, such as the current event, the circle of followers and followings, the verification status of the channel or user and clustering the events based on time and trends, may further refine and improve the performance of the model.

\newpage

\bibliographystyle{unsrtnat}
\bibliography{template}  %%% Uncomment this line and comment out the ``thebibliography'' section below to use the external .bib file (using bibtex) .





% \end{thebibliography}

\end{document}